\documentclass{article}

\usepackage[nonatbib]{arxiv}
\usepackage[T1]{fontenc}    
\usepackage{hyperref}       
\usepackage{url}            
\usepackage{booktabs}       
\usepackage{amsfonts}       
\usepackage{nicefrac}       
\usepackage{microtype}      
\usepackage{lipsum}

\usepackage{microtype}
\usepackage{graphicx}
\usepackage{subfigure}
\usepackage{booktabs} 

\usepackage{mathtools}

\DeclarePairedDelimiter\floor{\lfloor}{\rfloor}
\newcommand{\cropfn}[2]{f_{crop}^{{#1}\rightarrow{#2}}}
\newcommand{\padfn}[2]{f_{pad}^{{#1}\leftarrow{#2}}}
\newcommand{\distillfn}[2]{{Distill}^{{#1} \rightarrow {#2}}}
\newcommand{\distillsym}[2]{C{#1}\swarrow^{({#2})}}

\title{Progressive Label Distillation: Learning Input-Efficient Deep Neural Networks}

\author{
   Zhong Qiu Lin \thanks{
      Vision and Image Processing Research Group, Department of System Design Engineering, University of Waterloo, Waterloo, ON, Canada;
      DarwinAI Corp., Waterloo, ON, Canada;
   } \\
   University of Waterloo \\
   Waterloo, ON, Canada \\
   \And
   Alexander Wong \thanks{
      Vision and Image Processing Research Group, Department of System Design Engineering, University of Waterloo, Waterloo, ON, Canada;
      Waterloo Artificial Intelligence Institute, University of Waterloo, Waterloo, ON, Canada;
      DarwinAI Corp., Waterloo, ON, Canada
   } \\
   University of Waterloo \\
   Waterloo, ON, Canada \\
}

\begin{document}
\maketitle

\begin{abstract}
Much of the focus in the area of knowledge distillation has been on distilling knowledge from a larger teacher network to a smaller student network. However, there has been little research on how the concept of distillation can be leveraged to distill the knowledge encapsulated in the training data itself into a reduced form. In this study, we explore the concept of progressive label distillation, where we leverage a series of teacher-student network pairs to progressively generate distilled training data for learning deep neural networks with greatly reduced input dimensions. To investigate the efficacy of the proposed progressive label distillation approach, we experimented with learning a deep limited vocabulary speech recognition network based on generated 500ms input utterances distilled progressively from 1000ms source training data, and demonstrated a significant increase in test accuracy of almost 78\% compared to direct learning.
\end{abstract}

\section{Introduction}
\label{introduction}
Deep learning has been widely adapted to many different problems, such as image classification~\cite{krizhevsky2012imagenet}, speech recognition~\cite{hinton2012deep} and natural language processing~\cite{mikolov2010recurrent}, and has demonstrated state-of-the-art results for these problems. Despite the promises, deep neural networks (DNNs) remain challenging to deploy in on-device edge scenarios such as mobile and other consumer devices.

Due to the limited computational resources available in such on-device edge scenarios, many recent studies~\cite{sandler2018mobilenetv2, iandola2016squeezenet, lin2018edgespeechnets,ferminets} have put greater efforts into designing small, low-footprint deep neural network architectures that are more appropriate for embedded devices. A particularly interesting approach for enabling low-footprint deep neural network architectures is the concept of knowledge distillation~\cite{hinton2015distilling}, where the performance of a smaller network is significantly improved by leveraging a teacher-student strategy where the smaller network is trained to mimic the behaviour of a larger teacher network. With much of the research around distillation focused on  distilling knowledge from larger networks to smaller networks, there is little research focused on leveraging the concept of distillation for distilling knowledge encapsulated in the training data itself into a reduced form. By producing data with reduced data dimension, one can achieve input-efficient deep neural networks with significantly reduced computational costs.

In this study, we explore a concept we will call \textit{progressive label distillation}, where a series of teacher-student network pairs are leveraged to progressively generate distilled training data.  The proposed approach enables  the learning of computationally efficient DNNs with greatly reduced input dimensions without the need for collecting and labeling new data. The proposed strategy can be used in conjunction with any efficient deep neural network architecture to further reduce computational costs and memory footprint.

\section{Related Work}
\label{related work}
The proposed concept of label distillation stems from the notion of teacher-student learning, where a student network is learned using a teacher deep neural network or an ensemble of teacher deep neural networks.  A number of different teacher-student learning strategies have been explored in past literature~\cite{hinton2015distilling, ba2014deep, bucilu2006model, radosavovic2018data, Li2017}.  For example, in \cite{ba2014deep}, the authors explore the learning of a shallow student deep neural network to mimic the behaviour of a deeper teacher deep neural network. In \cite{hinton2015distilling}, the authors introduced the concept of a distillation loss for learning a student deep neural network from an ensemble of teacher deep neural networks. Both~\cite{ba2014deep} and \cite{hinton2015distilling} illustrate the promising results of leveraging teacher-student learning where the teacher-student pair have heterogeneous network architectures but leverage identical training data. Meanwhile, teacher-student learning  has also been adapted for the purpose of training data transformations, such as for the purpose of expanding the quantity of training data as well as for the purpose of domain adaptation. In \cite{bucilu2006model}, the authors explored using an ensemble of networks to generate synthetic data for learning shallow neural networks. In \cite{radosavovic2018data}, the authors demonstrated the concept of data distillation, which leverages a large quantity of unlabelled real data and ensembles the outputs of a neural network on multiple transformations of the unlabelled data for "self-training". In \cite{Li2017}, the authors leveraged teacher-student learning for transferring knowledge learned in the source data domain to the target data domain. All these related teacher-student approaches focus on exploiting deep neural networks with identical input dimensions using training data with fixed dimensions, as well as focusing on leveraging a one single step of teacher-student learning. As discussed previously, the notion of leveraging a series of teacher-student network pairs in a progressive manner for distilling the knowledge encapsulated in the training data itself into a reduced form has not been previously explored.

\section{Method}
\label{method}
The goal of the proposed \textit{progressive label distillation} approach is to distill knowledge from a teacher network with a large input dimension to a student network with a small input dimension, which in effect reduces computational costs significantly.  This is achieved by leveraging a series of teacher-student network pairs to progressively generate distilled data for training subsequent networks.

An overview of the progressive label distillation strategy can be described as follows. First, we train a teacher network using an original training data with dimensions greater than the target input dimension. Second, a new training data with reduced dimensions is generated from the original training data (e.g., in the case of limited vocabulary speech recognition, one can generate short audio samples by randomly cropping segments from longer audio samples), with the associated labels generated using the prediction results of the teacher network for these dimension-reduced samples. Since the labels are generated based on the knowledge of the teacher network, we will refer to these generated labels as \textit{distilled labels}. Third, a student network with reduced input dimensions is trained with the new input data and the distilled labels generated using the teacher network. This process is repeated in a progressive manner until the desired target input dimension is reached.  Details of these steps are described below.

\subsection{Label Distillation}
Let us first describe the proposed concept of label distillation.  Given a dataset $D = {\{(X_i, Y_i)\}}$, we wish to generate a new dataset $\Tilde{D} = {\{(f(X_i), Y_i)\}}$, where $f$ is a stochastic function that generates the new data based on a random process. In this study, the goal is to learn deep neural networks with reduced input dimensions without the need for collecting and labeling new training data, which in effect reduces computational costs significantly.

One can generate a new set of training data by randomly cropping the original training data to a target dimension, which is formally represented as ${\{\Tilde{X_i}\}} = {\{\cropfn{src}{tgt}(X_i)\}}$ where $\cropfn{src}{tgt}$ is a stochastic function that randomly crops from a source sample $X_i$ of dimension $src$ to produce a target sample $\Tilde{X_i}$ of dimension $tgt$. Thus, the dimension-reduced training data is $\Tilde{D} = {\{(\Tilde{X_i}, Y_i)\}} = {\{(\cropfn{src}{tgt}(X_i), Y_i)\}}$. However, leveraging the original class label of the original source sample $X_i$ directly as the class label of the target samples $\Tilde{X_i}$ derived from $X_i$ can lead to a poor reflection of the underlying content in $\Tilde{X_i}$.  For example, in the case of limited vocabulary speech recognition, a random crop from an audio sample with a label of 'yes' may actually contain no verbal utterance in it, and as such should really have a label of 'silence' assigned to it.  Similarly, in the case of image classification, random crops of an image labeled as 'desk' may contain predominantly a cup, a water bottle, a pen, a mouse, etc., and thus the label 'desk' is not reflective of these crops.

We address the aforementioned problem by generating new training data $\Tilde{D}$ by leveraging a teacher network trained on the original training data $D$ to generate \textit{distilled labels} for the reduced-dimension training data $\Tilde{D}$. More specifically, we modify $\{\Tilde{X_i}\}$ to match the input dimension of the teacher network (in this study, by padding the reduced-dimension training data $\{\Tilde{X_i}\}$).  Let us denote the padded reduced-dimension training data as $\{\Breve{X_i}\} = \{\padfn{src}{tgt}(\Tilde{X_i})\}$, where $\padfn{src}{tgt}$ is a deterministic function that pads $\Tilde{X_i}$ of size $tgt$ to the size of $src$. Hence, the newly generated training data $\Breve{D}$ for training a student network with reduced input dimensions can be defined as
\begin{equation}
    \label{eq:dataset}
    \begin{split}
        \Breve{D} & = \{ (\Tilde{X_i}, C^{src}(\Breve{X_i}) \} \\
                  & = \{( \cropfn{src}{tgt}(X_i), C^{src}(\padfn{src}{tgt}(\cropfn{src}{tgt}(X_i)) )\}
    \end{split}
\end{equation}

In this study, the teacher network $C^{src}$ is trained using the cross-entropy loss against the ground-truth labels $\{Y_i\}$ provided in the original training data $D$, with the student network $C^{tgt}$ for $tgt < src$ trained by minimizing the cross-entropy loss between its output and the distilled label generated by the teacher network $C^{src}$.

Letting $p^{src}_c(\Tilde{x}) \overset{\Delta}{=} C^{src}(\Breve{x})[c]$ be the probability assigned to class $c$ in the output of the teacher network $C^{src}$ on the padded dimension-reduced training data $\Breve{x}$, the loss function defined for the student network $C^{tgt}$ is formally represented in equation \ref{eq:loss}. It is worth noting that the student network is not exposed to the ground-truth labels $\{Y_i\}$ from the original training data.
\begin{equation}
    \label{eq:loss}
    L^{tgt}(\cropfn{src}{tgt}(X_i)) =
        -\sum_c p_c^{tgt}(\Tilde{X_i}) \log(p_c^{src}(\Breve{X_i}))
\end{equation}

We formally represent our label distillation as a function, shown in equation \ref{eq:distill}, that outputs a new student network $C^{tgt}$ with input dimension $tgt$, given a teacher network $C^{src}$ with input dimension $src$ and a dataset $D$.
\begin{equation}
    \label{eq:distill}
    C^{tgt} = \distillfn{src}{tgt}(D, C^{src})
\end{equation}

\paragraph{Hard Label vs. Soft Label}
Similar to \cite{hinton2015distilling}, one can train a network with soft labels by using the class probability output of the teacher network, $p^{src}_c(\Tilde{x}) \overset{\Delta}{=} C^{src}(\Breve{x})[c]$ as the label. We, thus, minimize the KL-divergence between the student network's output and the teacher network's output. On the other hand, we can also leverage the teacher network's output as the "ground-truth" hard label, and minimize the cross-entropy loss between the student model's output and the mutual exclusive label, ${argmax}_{\forall c \in \mathcal{C}}( C^{src}(\Breve{x})[c])$.  In this study, we evaluate the efficacy of both a hard label and a soft label approach for label distillation.

\subsection{Progressive Label Distillation}
For the purpose of learning an input-efficient deep neural network with an input dimension of $tgt$, the proposed concept of label distillation can be leveraged in two different ways: i) direct, and ii) progressive.  In the direct label distillation approach, a teacher network $C^{src}$ is used to produce distilled labels for a newly generated training data of dimension $tgt$, and a student network $C^{tgt}$ is then trained on the newly generated training data and the corresponding distilled labels.  This direct approach is encapsulated by $\distillfn{src}{tgt}$ (Eq.~\ref{eq:distill}).

In the progressive label distillation approach, a series of teacher-student network pairs are leveraged to progressively generate distilled labels for a series of generated training data with progressively lower dimensions until the target dimension $tgt$ is reached.  Furthermore, an interesting by-product of this progressive label distillation process is a series of student networks with progressively lower input dimensions, with the final student network being an input-efficient deep neural network possessing the target input dimension.

Letting $int\in(tgt, src)$ denote a number of intermediate dimensions between $tgt$ and $src$, the progressive label distillation process can be expressed by
\begin{equation}
    C^{tgt} = \distillfn{int}{tgt}(D, \distillfn{src}{int}(D, C^{src}))
\end{equation}

For illustration purposes, we define a convenient notation for representing the student networks produced via label distillation.  We denote a student network with an input dimension of $tgt$ that is learnt via direct label distillation from a teacher network with an input dimension of $src$ as $\distillsym{tgt}{src}$. In addition, we denote a student network with an input dimension of $tgt$ that is learnt via progressive label distillation from a teacher networks with input dimensions of $int$ and $src$ as $\distillsym{tgt}{int,src}$.

\begin{figure*}
    \centering
    \includegraphics[width=\textwidth]{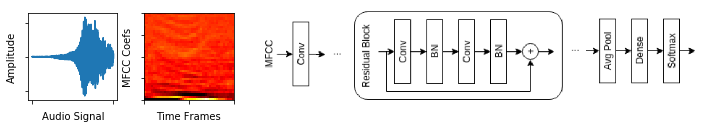}
    \caption{(left) Audio signal; (middle) MFCC representation; (right) deep residual network architecture (res15) proposed in{~\cite{tang2018deep}}. }
    \label{fig:speech_architecture}
\end{figure*}

\section{Experiments and Discussion}
\label{experiment}
To better investigate and explore the efficacy of the introduced notions of label distillation and progressive label distillation, a number of experiments were performed for the task of limited vocabulary speech recognition~\cite{sainath2015convolutional,tang2018deep,lin2018edgespeechnets}, where the underlying goal is to identify which word from a limited vocabulary was spoken based on an input audio utterance recording.  In general, we will first explore direct label distillation for learning student networks with various input dimensions.  We will then investigate the effectiveness of progressive label distillation through different teacher-student network pair configurations.

\subsection{Experimental Setup}
As mentioned previously, we will leverage the task of limited vocabulary speech recognition to investigate the efficacy of label distillation for producing student input-efficient deep neural networks from teacher networks with larger input dimension.  Here, we will describe the experimental setup used in this study.

\subsubsection{Dataset}
In this study, the Speech Command dataset~\cite{warden2018speech_command} was used for experimental purposes.  The Speech Command dataset consists of 105,829 one-second utterances of 35 different words spoken by 2618 speakers, categorized into 12 different classes: silence, an unknown word, "yes", "no", "up", "down", "left", "right", "on", "off", "stop", and "go".

\subsubsection{Feature Extraction and Input Pre-processing}
\label{feature_extraction}
Based on the past literature \cite{hinton2012deep}, a very effective strategy for leveraging deep neural networks for limited vocabulary speech recognition is to first transform the input audio signal into mel-frequency cepstrum coefficient (MFCC) representations (see Figure \ref{fig:speech_architecture}). For reducing audio signal noise, a band-pass filter of 20Hz/4kHz is applied to the input audio. Inspired by \cite{sainath2015convolutional}, the input feature is forty dimensional MFCC frames stacked using a 30ms window and 10ms frame shift. Substantially, the MFCC representation of the audio signal is used as the input to the deep neural network.

\subsubsection{Model Architecture}
\label{model architecture}
In this study, we leveraged the deep residual network architecture proposed by  \cite{tang2018deep}, which they refer to as res15 and was shown to provide state-of-the-art accuracy when it was first published. In particular, Tang et al. proposed to use a residual block architecture where the first layer of the block is a bias-free convolutional layer with weights $W\in \mathbb{R}^{(m\times r)\times (n_{i-1}\times n_i)}$, where $m$ and $r$ are the width and height of convolutional kernel, and $n_{i-1}$ and $n_i$ are the number of channels for the previous convolutional layer and the current convolutional layer, respectively. After the convolutional layer, a ReLU activation and  batch normalization~\cite{ioffe2015batch} is appended in the residual block. In addition, convolutional dilation, $(d_w, d_h)$, is used to increase the receptive field of the network. Increasing reception fields in deeper layers allow the network to consider the input entirely without the need of very deep layers.

The details of the network architecture is shown in table \ref{table1res15}, and Fig.~\ref{fig:speech_architecture} (right) shows the overall architecture and the detail of one of the residual blocks.  For the teacher network, we constructed a network based on this network architecture with an input dimension for an 1000ms MFCC representation of the audio signal.  To explore direct and progressive label distillation strategies for learning student networks with various input dimensions, we modify the input layer to construct a series of input-efficient networks with input dimension for MFCC representations of audio signals with the following lengths: 900ms, 800ms, 700ms, 600ms, and 500ms.

\begin{table}[]
\centering
    \caption{res15 network architecture proposed in \cite{tang2018deep}}
    \label{table1res15}
    \begin{tabular}{l|lllll}
        Type          & $m$        & $r$        & $n$        & $d_w$                      & $d_h$                      \\ \hline
        conv          & 3          & 3          & 45         & -                          & -                          \\
        res$\times$6  & 3          & 3          & 45         & $2^{\floor{\frac{i}{3}}}$  & $2^{\floor{\frac{i}{3}}}$  \\
        conv          & 3          & 3          & 45         & 16                         & 16                         \\
        bn            & -          & -          & 45         & -                          & -                          \\
        avg-pool      & -          & -          & 45         & -                          & -                          \\
        softmax       & -          & -          & 12         & -                          & -
    \end{tabular}
\end{table}
\begin{table}[]
\centering
\caption{Accuracy of the student networks produced using direct distillation.  Computed based on the average predictions of three random crops.}
\label{table:soft-hard-gt}
\begin{tabular}{c|cc}
     & \multicolumn{2}{c}{Test Acc (\%)} \\
        & \multicolumn{1}{c}{soft}
        & \multicolumn{1}{c}{hard} \\ \hline
$C1000$ & N/A & 95.8                 \\
$C500$ & N/A & 12.03                 \\\hline
$\distillsym{500}{1000}$ & 85.82                   & 83.81                                   \\
$\distillsym{600}{1000}$       &  90.12                   & 89.79                                   \\
$\distillsym{700}{1000}$        & 92.58                   & 92.36                               \\
$\distillsym{800}{1000}$ & 93.81                   & 94.01                                      \\
$\distillsym{900}{1000}$ & 94.91                   & 95.23
\end{tabular}
\end{table}

\subsubsection{Training}
All networks used in this study were trained using TensorFlow~\cite{abadi2016tensorflow} on a single Nvidia GTX 1080 Ti GPU with batch size of $64$ for $25$ epochs to ensure convergence.  The number of random crops used to train the student networks are identical for all tests in the study.  The learning rate is constant with the initial learning being $0.1$ for the first $6000$ iterations. It is divided by $10$ after $6000$ iterations, and again divided by $10$ after $12000$ iterations. We use SGD with momentum of $0.9$, and the decay steps and decay rate are set to $1000$ and $0.9$, respectively. Weight decay regularization is also imposed with strength of $0.00001$.

\subsection{Performance Evaluation}
As the input dimension of the student networks in this study does not match the dimensions of the original dataset $D$, it is no longer feasible to evaluate the accuracy of these networks using the test dataset in its original form. To facilitate for a fair and consistent evaluation of the learnt student networks with differing input dimensions, three random crops with dimensions equal to that of the student network being tested are extracted from each 1000ms audio sample in the test dataset, and passed into the student network to obtain prediction outputs.   Given the predictions made by the student network on the three random crops, a final prediction is calculated as the average of three predictions. This final prediction is evaluated against the ground-truth label for assessing the accuracy of the student network.

The offset values used to extract random crops from the original sample follow a uniform distribution between zero and the difference between the original test sample length and the input dimension of the student network, which can be expressed as $\{T_{offset}\} = \{ \floor{\frac{0.5}{3}(src-tgt)}, \floor{\frac{1.5}{3}(src-tgt)}, \floor{\frac{2.5}{3}(src-tgt)} \}$.

\subsection{Experiment 1: Direct Label Distillation}
\begin{figure*}
    \centering
    \includegraphics[width=0.7\textwidth, height=0.4\textwidth]{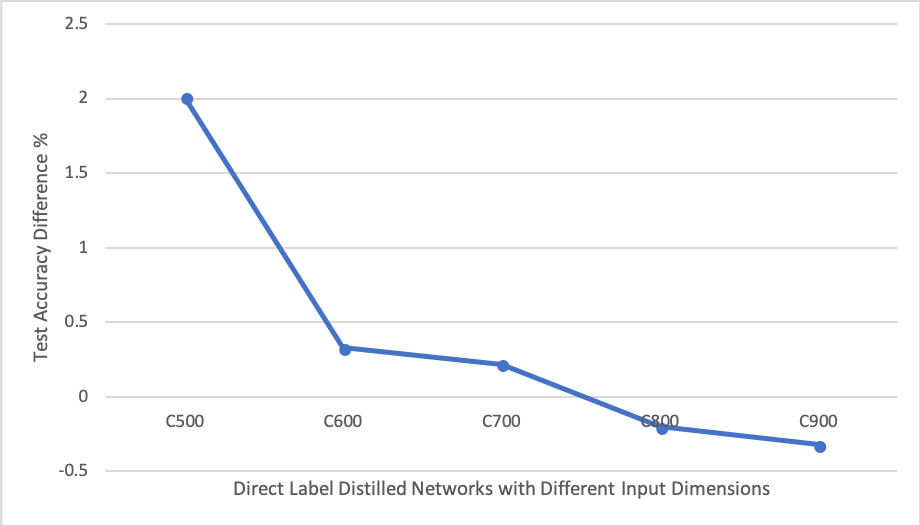}
    \caption{Test accuracy difference between using soft labels and hard labels for label distillation. It can be observed that the test accuracy difference increases monotonically as the target input dimension decreases, with the biggest jump in test accuracy difference between using soft labels and hard labels occurring when the target input dimension decreases from 600ms to 500ms (i.e., $\distillsym{600}{1000}$ vs. $\distillsym{500}{1000}$).}
    \label{fig:soft_hard_diff}
\end{figure*}
In the first experiment, we evaluate the efficacy of direct label distillation of learning input-efficient student networks via soft labels and hard labels for a set of student networks with five different input dimensions (i.e., $\{500ms, 600ms, 700ms, 800ms, 900ms\}$) by computing their respective test accuracies.  For comparison purposes, we also evaluate the test accuracy of a student network with input dimension corresponding to MFCC representations of a 500ms audio signal learnt directly from random crops without label distillation (referred to here as C500).  Finally, as a baseline reference, the test accuracy of the teacher network with full-length input dimension (referred to here as C1000) is also evaluated.

The test accuracies for the aforementioned student networks learned via direct label distillation, along with C500 and C1000, are shown in Table~\ref{table:soft-hard-gt}.  A number of interesting observations can be made.  First, it can be observed that the network with input dimension of 500ms trained directly from random crops without label distillation (i.e., C500) was able to achieve a test accuracy of just 12.03\%, while the student networks with input dimension of 500ms (i.e.,  $\distillsym{500}{1000}$) learnt using label distillation via soft labels and hard labels was able to achieve test accuracies of 85.82\% and 83.81\%, respectively.  The significant increase in test accuracy when using direct label distillation compared to direct learning clearly illustrates the effectiveness of label distillation for learning input-efficient deep neural networks.

Secondly, it can be observed that while the use of hard labels for direct label distillation achieves higher accuracies than the use of soft labels for student networks with larger input dimensions, the opposite is true for student networks with smaller input dimensions.  This is more evident when we study the test accuracy difference between the use of soft labels and hard labels (see Figure \ref{fig:soft_hard_diff}), where the test accuracy difference increasing gradually as the target input dimension decreases, with the biggest jump in test accuracy difference between using soft labels and hard labels occurring when the target input dimension decreases from 600ms to 500ms (i.e., $\distillsym{600}{1000}$ vs. $\distillsym{500}{1000}$).  As such, the use of soft labels in label distillation improves the test accuracy over using hard labels when the difference between the source input dimension $src$ of the teacher network and the target input dimension $tgt$ of the student network is larger. Intuitively, the use of soft labels provides more information on the shared characteristics between classes, and this additional "soft" class information can help the student network learn when the input information is more limited.

\subsection{Experiment 2: Progressive Label Distillation}
\begin{table}[]
\centering
\caption{Test accuracy of networks learnt using progressive label distillation: each column shows the test accuracy of networks that have the same input dimension.  Each row shows the number of progressive label distillation steps. Each network in the occupied cell is learnt via label distillation from the network in the first occupied cell to its left.}
\label{table:progressive_label_distillation}
\begin{tabular}{c|rrrrr}
\# Steps & \multicolumn{1}{c}{C900} & \multicolumn{1}{c}{C800} & \multicolumn{1}{c}{C700} & \multicolumn{1}{c}{C600} & \multicolumn{1}{c}{C500} \\ \hline
1     &                         &                         &                         &                         & 85.82                  \\ \cline{1-1}
2     &                         &                         &                         & 90.12                  & 86.71                  \\
2     &                         &                         & 92.58                  &                         & 87.92                  \\
2     &                         & 93.81                  &                         &                         & {\textbf{89.22}}       \\
2     & 94.91                  &                         &                         &                         & 88.94                  \\ \cline{1-1}
3     &                         &                         & 92.58                  & 91.18                  & 84.98                  \\
3     &                         & 93.81                  &                         & 90.97                  & 88.24                  \\
3     & 94.91                  &                         & 93.07                  &                         & 86.12                  \\ \cline{1-1}
4     &                         & 93.81                  & 93.38                  & 90.96                  & 84.90                  \\
4     & 94.91                  & 94.34                  & 92.85                  &                         & 84.25                  \\
 \cline{1-1}
5     & 94.91                  & 94.34                  & 92.85                  & 89.15                  & 79.50
\end{tabular}
\end{table}

\begin{figure*}
    \centering
    \includegraphics[width=0.7\textwidth, height=0.4\textwidth]{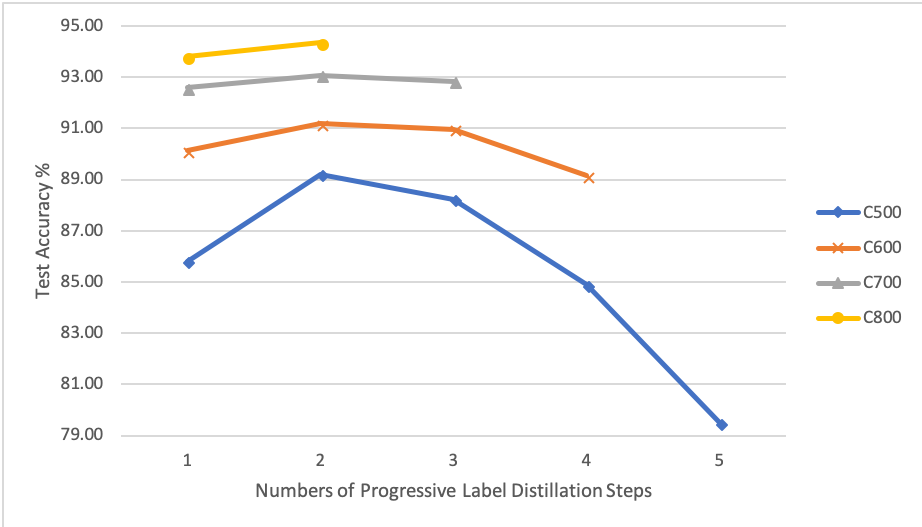}
    \caption{Impact on test accuracy for different number of steps of progressive label distillation}
    \label{fig:progresivs_label_distillation}
\end{figure*}

In the second experiment, we evaluate the efficacy of progressive label distillation of learning input-efficient student networks for different combinations of series of teacher-student networks with progressively smaller input dimensions.  More specifically, we evaluate the test accuracies achieved by the student networks without a progressive label distillation framework with different number of distillation steps, $\{\# Steps\} = \{1, 2, 3, 4, 5\}$. The test accuracy of the student networks learnt using progressive label distillation is shown in Table~\ref{fig:progresivs_label_distillation}.  Each network in the occupied cell of Table~\ref{fig:progresivs_label_distillation} is learnt using distilled labels generated by the network in the first occupied cell to its left. For example, the student network with 500ms input dimension with a test accuracy of $89.22\%$ (shown in bold) (i.e., $\distillsym{500}{800, 1000}$), is learnt using distilled labels generated by a network with 800ms input dimension, which itself was learnt using distilled labels generated by a network with 1000ms input dimension.  By performing label distillation in a progressive manner, the student network $\distillsym{500}{800, 1000}$ achieves a significant boost in test accuracy of $+3.40\%$ when compared to the student network $\distillsym{500}{1000}$ which was learnt via direct label distillation. This improvement in test accuracy is not unique to $\distillsym{500}{800, 1000}$, as progressive label distillation improves test accuracy for a wide range of different teacher-student configurations when compared to direct label distillation. Figure \ref{fig:progresivs_label_distillation} shows that for all target dimensions ${tgt} = \{500, 600, 700, 800\}$, progressive label distillation achieved higher test accuracy when compared to direct label distillation.

While significant increases in accuracy was observed when using the proposed progressive label distillation approach to learn input-efficient networks, it can also be observed that too many steps for progressive label distillation does not yield better test accuracy. More specifically, as shown in Figure~\ref{fig:progresivs_label_distillation}, while two-step progressive label distillation improved the performance across all target input dimensions, the test accuracy of student networks starts to drop as more steps are leveraged for progressive label distillation. In fact, four-step and five-step progressive label distillation results in test accuracies below that of direct label distillation.  In addition, we observed that what intermediate teacher-student network pair configuration is chosen has noticeable impact on the final target student network. As such, it is very much worth further investigation on how best to determine the appropriate series of  teacher-student network pairs for optimal performance.

\subsection{Computational Cost of Input-Efficient Networks}
With reduced input dimension, the computational cost of the student networks are greatly reduced. Table \ref{table:flops} shows the number of FLOPs of different operations for the student networks with different input dimensions. It can be observed that the number of FLOPs for $C500$ is less than half of the number of FLOPs for $C1000$.  As such, the use of progressive label distillation can allow for the creation of input-efficient networks with low computational costs while still providing reasonable test accuracies.

\begin{table}[]
\centering
\caption{FLOPs for input-efficient networks with different input dimensions. The teacher network (C1000) is provided as a reference for comparison.}
\label{table:flops}
\begin{tabular}{c|rrrrr}
      {Model} & \multicolumn{1}{c}{Conv} & \multicolumn{1}{c}{Add} & \multicolumn{1}{c}{Pool} & \multicolumn{1}{c}{Total} & \multicolumn{1}{c}{Factor} \\ \hline
C500  & 775M    & 472k    & 3.78k    & 776M    & 0.47  \\
C600  & 951M    & 575k    & 6.48k   & 952M    & 0.58  \\
C700  & 1.13B   & 677k    & 9.18k    & 1.13B   & 0.69  \\
C800  & 1.29B   & 780k    & 11.9k   & 1.29B   & 0.79  \\
C900  & 1.47B   & 882k    & 14.6k   & 1.47B   & 0.90  \\
C1000 & 1.64B   & 985k    & 17.3k   & 1.64B   & 1.00
\end{tabular}
\end{table}

\section{Conclusions and Future Work}
\label{conclusion}
In this study, we show that progressive label distillation can be leveraged for learning deep neural networks with reduced input dimensions without collecting and labeling new data.  This reduction in input dimension results in input-efficient networks with significant reduction in the computation cost.  Experiment results for the task of limited vocabulary speech recognition show that the use of progressive label distillation can lead to an input-efficient student network with half the input dimension with a test accuracy of $89.22\%$, compared to just $12.03\%$ without using label distillation.

Future work involves investigate the effectiveness of progressive distillation for a broader range of tasks such as video action recognition and video pose estimation.  Furthermore, it would be worth exploring and expanding progressive label distillation to teacher-student network pairs with heterogeneous network architectures.

\section*{Acknowledgment}
\label{ack}
The authors would like to thank the Natural Sciences and Engineering Research Council of Canada, the Canada Research Chairs program, and DarwinAI Corp.

\bibliographystyle{unsrt}
\bibliography{references}

\end{document}